\documentclass[10pt, a4paper]{article}
\usepackage{lrec}
\usepackage{multibib}
\newcites{languageresource}{Language Resources}
\usepackage{graphicx}
\usepackage{tabularx}
\usepackage{soul}
\usepackage{xcolor}

\usepackage{footnote}
\usepackage{xstring}
\usepackage{booktabs}
\usepackage{wrapfig}
\usepackage{graphicx,tikz}
\usepackage{subcaption,multirow}
\usepackage{amsmath,color}
\usepackage{tablefootnote}
\usepackage{threeparttable}
\usepackage{amsfonts}
\usepackage{multicol}
\usepackage{tablefootnote}
\usepackage{pgfplots}
\usepackage{verbatim}
\usepackage{comment}
\usepackage{enumitem} 
\usepackage{linguex} 
\usepackage{cognac_v1.0.3}

\usepackage{titlesec}
\titleformat{\section}{\normalfont\large\bf\center}{\thesection.}{1em}{}
\titleformat{\subsection}{\normalfont\SmallTitleFont\bf\raggedright}{\thesubsection.}{1em}{}
\titleformat{\subsubsection}{\normalfont\normalsize\bf\raggedright}{\thesubsubsection.}{1em}{}
\renewcommand\thesection{\arabic{section}}
\renewcommand\thesubsection{\thesection.\arabic{subsection}}
\renewcommand\thesubsubsection{\thesubsection.\arabic{subsubsection}}

\usepackage{epstopdf}
\usepackage[utf8]{inputenc}

\usepackage{hyperref}
\usepackage{xstring}

\usepackage{color}

\newcommand{\codecite}{Data and code may be downloaded from \url{https://doi.org/10.34703/gzx1-9v95/WLHYTR}.}

\newcommand{\emo}[2]{[#1]\textsuperscript{#2}}

\title{ \vspace*{.5\baselineskip} \bf The ABBE Corpus: Animate Beings Being Emotional}

\name{Samira Zad, Joshuan Jimenez, \& Mark A. Finlayson}

\address{Knight Foundation School of Computing and Information Sciences \\
Florida International University \\ 
11200 S.W. 8th St., Miami, FL\\
\texttt{\{szad001, jjime178, markaf\}@fiu.edu}}

\abstract{Emotion detection is an established NLP task of demonstrated utility for text understanding. However, basic emotion detection leaves out key information, namely, {\it who} is experiencing the emotion in question. For example, it may be the author, the narrator, or a character; or the emotion may correspond to something the audience is supposed to feel, or even be unattributable to a specific being, e.g., when emotions are being discussed {\it per se}. We provide the ABBE corpus---Animate Beings Being Emotional---a new double-annotated corpus of texts that captures this key information for one class of emotion experiencer, namely, animate beings in the world described by the text. Such a corpus is useful for developing systems that seek to model or understand this specific type of expressed emotion. Our corpus contains 30 chapters, comprising 134,513 words, drawn from the Corpus of English Novels, and contains 2,010 unique emotion expressions attributable to 2,227 animate beings. The emotion expressions are categorized according to Plutchik's 8-category emotion model, and the overall inter-annotator agreement for the annotations was 0.83 Cohen's Kappa ($\kappa$), indicating excellent agreement. We describe in detail our annotation scheme and procedure, and also release the corpus for use by other researchers\footnote{\codecite}.\\ \newline  }
\pgfplotsset{compat=1.17}
\begin{document}

\maketitleabstract

\section{Introduction}

Emotion detection is an NLP task that has been established for quite some time, and has seen quite a few published corpora, resources, and systems \cite{olveres1998intelligent,mueller1998natural,aman2008using,chatterjee2019semeval,zad2020systematic,zad2021survey}. Traditionally, emotion detection consists of categorizing a piece of text as to an expressed emotion, for example, tagging the sentence ``I was furious.'' with the label {\sc anger}. Alternatively, the task might also involve first finding spans of text that express emotion before categorizing them, for example identifying that it is the word {\it furious} that provides the affective semantics for that sentence. 

This approach to emotion detection is useful, but notably leaves out a key piece of information: namely, who exactly is experiencing the emotion. Emotions usually do not appear in a vacuum, and are usually experienced by {\it someone}, and knowing who is experiencing the emotion is a important step in understanding the semantics of the text. Accordingly we present the first corpus where emotion expressions are associated with those who are experiencing the emotion. In the general case, emotions might be associated with several different types of experiencers (e.g., the author, the narrator, the audience, etc.), each presenting their own challenges for definition and annotation. In the ABBE corpus we focus on animate beings who are part of the world of the text and are experiencing emotions, and we provide annotations on top of 30 chapters (i.e., narratives) drawn from the Corpus of English Novels, providing several thousand animate beings and expressed emotions that can be used for training, testing, and validation of automatic systems.

The paper proceeds as follows. First, we review basic definitions of the key concepts in play, i.e., {\it emotion} and {\it animate being} (\S\ref{sec:definitions}). We then review prior work on the emotion theory we use, emotion detection systems, annotation schemes for emotion and animate beings, and existing corpora for both (\S\ref{sec:priorwork}). We then describe in detail the annotation scheme we designed (\S\ref{sec:scheme}), as well as the texts included in the corpus, our selection criteria, and agreement measures (\S\ref{sec:texts}). We conclude with an examination of interesting and difficult edge cases (\S\ref{sec:edgecases}), and our contributions (\S\ref{sec:contributions})\footnote{The dataset can be downloaded from \url{https://doi.org/10.34703/gzx1-9v95/WLHYTR}}. 


\section{Definitions}
\label{sec:definitions}

\subsection{Emotion}

There are many ways of defining emotion, not all of which are relevant to the task of finding emotions in text. For example, we know of theories of emotion that go back to the Ancient Greeks and Romans---such as Aristotle, Cicero, Senaca, and Galen---and emotion remained a topic of theorizing through the Middle Ages (Augustine, Aquinas) and Renaissance (Machiavelli, Montaigne) \cite{sep-emotions-17th18th}. In the dawn of the scientific age of psychology, thinkers as august as Charles Darwin and William James found emotion to be worthy of their attention and effort \cite{darwin1872expression,james1894discussion}.

Modern theories of emotion have three main dimensions of explanation or description: physiological, neurological, and cognitive. According to physiological views, emotions are responses within the human body to external or internal stimuli. According to neuroscientific views, emotional reactions can be explained by neural processes in the brain.

Cognitive approaches, pursued in psychology and cognitive science, have generally been considered the most useful for text processing. The American Psychological Association (APA), for example, defines emotion as {\it a complicated reaction pattern that can be noted in various ways, where emotion is composed of elements such as behavioral, physiological, and experiential based on how an individual deals with an event that has significance to them}  \cite{vandenbos2007apa}. The Dictionary of Cognitive Psychology \cite{eysenck1994blackwell}, on the other hand, does not formally define emotion, but an operative definition emerges from its five pages devoted to emotion: emotion is a {\it mental state}. Cognitive theories of emotion vary in their complexity, with some theories identifying sophisticated constellations of components, including the activation of appraisals, the holding of subsequent desires, and the formation of intentions \cite{izard1992basic}. For some theorists, all cognition participates more or less in emotion \cite{scherer1993neuroscience}. Despite this range of complexity, what is critical for our work is that emotion is a mental state that must be attributed to a being capable of maintaining such a state. 

In cognitive / psychological approaches (which, as mentioned, foregrounds mental state) there are three broad classes of theories that attempt to describe what emotions exist and the interactions between them: categorical, dimensional, and hybrid. {\bf Categorical} models propose a discrete set of emotions; these include theories by \newcite{ekman1999basic}, \newcite{parrott2001emotions}, \newcite{shaver1987emotion}, \newcite{oatley1987towards}, and \newcite{izard2007basic}. {\bf Dimensional} theories propose descriptive dimensions of and relations between emotions, such that experienced or expressed emotions fall along the relevant dimensions and potentially shade into each other, and are not necessarily distinct. Theories in this class include those by \newcite{russell1999core}, \newcite{scherer2005emotions}, \newcite{lovheim2012new}, \newcite{ortony1990cognitive}, and \newcite{fontaine2007world}. Finally, there are {\bf Hybrid} models which combine aspects of both categorical and dimensional theories; the theory by Plutchik \cite{plutchik2001nature} falls into this class. In our work, for reasons given in the literature review section below, we adopt the eight basic categories of Plutchik's model. Regardless of the specific theory chosen, however, what remains is the importance of the concept of mental state, which means that a proper description of emotion includes not only the emotion itself, but also the experiencer of an emotion. We turn to that next.








\subsection{Experiencers of an Emotion: Animate Beings}   

Key to emotion is mental state, and to have a mental state there must be a mind. There are numerous minds---real or imagined---that could be the experiencer of emotions described in text. From the point of view of a reader of a text, possibly the first being that comes to mind as potentially experiencing an emotion is the {\bf reader} themself: individuals who consume a text, be it a narrative, essay, textbook, or other genre of text, can experience emotions during that consumption. These emotional experiences may or may not correspond to emotions described in the text---for example, a student might feel despair or anguish on reading the first paragraphs of their new textbook on statistical mechanics, despite those emotions not being directly describe in the textbook in question. There, of course, are notable cases where emotions described in a text might reasonably be expected to be experienced by the reader, such as the case in literature when readers feel sympathetic emotion that mirrors that being experience by characters in the narrative. This is obviously a difficult case because how an individual or an ``average reader'' (if such a person can even be reasonably constructed) might react to a particular text can be extremely difficult to predict.

Another possible emotion experiencer is the {\bf author} (as distinct from the presented narrator) of the text. In these cases, especially in cases of first person description, the author may describe or imply emotions attributable to themself. This can be communicated explicitly through semantically emotional words (e.g., {\it When writing this book, I fell into depression.}) or may be expressed implicitly through devices such as style or even punctuation (e.g., the use of explanation points).

A third broad category of experiencer we will term {\bf animate beings}, by which we specifically mean beings described as being part of the ``world'' of the text. Canonically, such beings are often thought of as the ``characters'' in the world of the text; however, our category is broader than character because there might beings described in the text which can experience emotions which are not, narratively speaking, characters. We thus more precisely define this concept.

We start by defining {\it animacy}, which is the characteristic of being able to independently carry out actions (e.g., movement, communication, etc.)~\cite{jahan2018animacy}. Human beings, for example, are animate because they can move and communicate in a realistic environment; however, a chair or a table cannot do these things on their own, hence they are typically regarded inanimate. Animacy is a required property of characters in stories, which means that all characters must be animate in the traditional sense.

Characters, however, are not the only possible animate beings in a text. As defined in detail elsewhere \cite{jahan2020character}, characters are {\it animate beings that are important to the plot of a narrative}, meaning that they have a non-trivial role in advancing the action described in the text. With this distinction in mind, one can see that not all beings mentioned in a text are necessarily character: the text in question might not actually be a proper narrative, or there might be other, minor beings that could in theory be removed while keeping the essence of the plot or action of the text. We include these beings in our definition of {\it animate being}. An animate being, then, is any entity described in a text that can act autonomously or individually such as a person, an animal, the narrator, an imaginary creature, a magically animate tree, etc. Any and all of these beings can potentially experience emotions in the text.

The challenges of identifying the emotions experienced by these three classes of potential experiencers of emotion (readers, authors, and animate beings) differs greatly; therefore, we concentrate in this work on the third class (animate beings), as these seem to account for the vast majority of explicit references to emotion in texts commonly encountered.

\section{Literature Review}
\label{sec:priorwork}

Now that we have defined our two main concepts for the ABBE corpus, we proceed to review key pieces of prior work.

\subsection{Plutchik’s Wheel of Emotions Model}

As mentioned above, there are numerous psychological theories of emotion. In this work we have chosen to use the eight basic emotions of Plutchik's model. The reasons for this choice are twofold. First, the most commonly used emotion theory in NLP is Ekman's six category model~\cite{ekman1999basic}; however, this model has several noted deficiencies including lacking several key emotional concepts (i.e., trust, anticipation) as well as lacking any well defined relationship between the categories. Plutchik's model resolves these two problems without adding a significant amount of complexity (which would increase the difficult of annotation), while still being compatible with Ekman's model. Second, there are a number of available resources for Plutchik's model, including both emotion lexicons and corpora, which make the choice of this as a model practical in terms of building immediately useful systems.

\begin{figure}
\centering
	    \includegraphics[width=0.8\columnwidth]{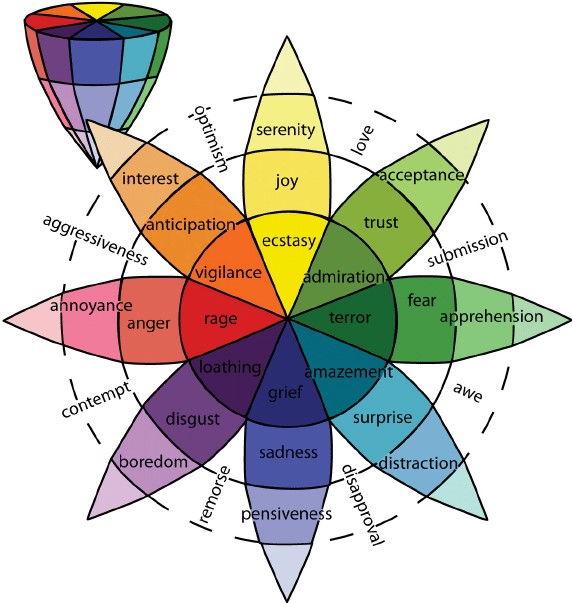}
	    \caption{Plutchik's emotions wheel, \protect\cite{plutchik1997circumplex}. Figure taken from \protect\cite{maupome2013dental}, with permission.}
	    \label{fig:plutchik}
\end{figure}

To understand Plutchik’s model, shown in Figure~\ref{fig:plutchik}, it is necessary to break it down based on the various aspects of the model, such as what the primary emotions are and their opposites.

\paragraph{Primary} The eight colored sections are designed to indicate that there are eight primary emotions: anger, anticipation, joy, trust, fear, surprise, sadness and disgust. Some attributes that are associated with these eight sectors can be seen in Table~\ref{tab:PrimaryRelated}. The eight primary emotions can be seen related to certain scenarios and cognition that help others understand what emotion an individual is experiencing in each moment and possible reasons for it. A person who feels as if their life is threatened, as shown in Table~\ref{tab:PrimaryRelated}, would most likely feel fear of dying and try to leave the situation to feel safe again. Throughout this, normally a person would feel fear and would express such emotion. This is further supported in Table~\ref{tab:PrimaryHypothesized} as shown with certain cases such as reacting to contact with a strange object can surprise people. This can be seen too with anger when someone is destructive. Emotions can be shown through the actions, words, responses, and more from an individual in a certain situation.

\paragraph{Opposite} Each primary emotion has a polar opposite:

\begin{itemize}[nosep]
    \item Joy $\longleftrightarrow$ Sadness
    \item Fear $\longleftrightarrow$ Anger
    \item Anticipation $\longleftrightarrow$ Surprise
    \item Disgust $\longleftrightarrow$ Trust
\end{itemize}

\paragraph{Combinations} Emotions are often complex. The emotions placed between the colored sections are those represented as a mix of the two neighboring primary emotions. For example, Anticipation and Joy combine into Optimism. Joy and Trust combine into Love. 

\paragraph{Intensity} The vertical dimension, shown radially in the main portion of the figure, and vertically in the upper left inset, Figure~\ref{fig:plutchik}, represents intensity---emotions intensify as they move from the outside to the center of the wheel (top to bottom of the cone), which is also indicated by the color: The darker the shade, the more intense the emotion. For example, Anger at its lowest level of intensity is Annoyance. At its highest level of intensity, Anger becomes rage. Another is a feeling of Boredom, which can intensify to loathing if left unchecked, noted as dark purple. 

\begin{table*}[t]
\centering
\begin{tabular}{lllll}
\toprule
Stimulus Event          & Cognition   & Feeling State                & Overt Behavior     & Effect                   \\
\midrule  
Threat                  & Danger      & Fear                         & Escape             & Safety                   \\
Obstacle                & Enemy       & Anger                        & Attack             & Destroy Obstacle         \\

Gain   of valued object & Possess     & Joy                          & Retain   or repeat & Gain   resources         \\
Loss of valued object   & Abandonment & Sadness                      & Cry                & Reattach to lost object  \\

Member   of one’s group & Friend      & Acceptance   (Trust)         & Groom              & Mutual   support         \\
Unpalatable object      & Poison      & Disgust                      & Vomit              & Eject poison             \\

New   territory         & Examine     & Expectation   (Anticipation) & Map                & Knowledge   of territory \\
Unexpected event        & What is it? & Surprise                     & Stop               & Gain time to orient \\    
\bottomrule
\end{tabular}
\caption{Related attributes of feeling states \protect\cite{plutchik2001integration}.}
\label{tab:PrimaryRelated}
\end{table*}
\begin{table*}[t]
\centering
\begin{tabular}{ll}
\toprule
Prototype Adaptation    &	Hypothesized Emotion\\
\midrule  
Protection: Withdrawal, retreat, contraction &	Fear, Terror\\
Destruction: Elimination of barriers to the satisfaction of needs&	Anger, Rage\\
Incorporation: Ingesting nourishment&	Acceptance (Trust)\\
Rejection: Riddance response to harmful material&	Disgust\\
Reproduction: Approach, contact, genetic exchanges&		Joy, Pleasure\\
Reintegration: Reaction to loss of a nutrient object&	Sadness, Grief\\
Exploration: Investigation of one’s environment	& Curiosity, Play (Anticipation)\\
Orientation: Reaction to contact with a strange object&	Surprise\\
\bottomrule
\end{tabular}
\caption{Hypothesized Emotions \protect\cite{plutchik2001integration}.}
\label{tab:PrimaryHypothesized}
\end{table*}

\subsection{Language Resources for Emotion}

We review here a selection of language resources that are relevant to emotion, and provide context for the ABBE corpus. We review both emotion lexicons and annotated corpora.

\subsubsection{NRC \& Revised NRC Emotion Lexicons}

The NRC (National Research Council of Canada) or Emolex emotion lexicon \cite{mohammad2013nrc,mohammad2013crowdsourcing}, is a ``word-sense level'' emotion lexicon with 14,182 entries. It is a General Purpose Emotion Lexicon (GPEL) derived from widely available sources and applicable to all domains \cite{zad2021emotion}, and manually annotated through Amazon’s Mechanical Turk service into Plutchik's eight basic emotions. \newcite{zad2021hell} semi-automatically corrected and updated the NRC emotion lexicon by assigning POS tags to entries, manually reviewing problem entries, and matching NRC entries with other emotion lexicons to deduce label accuracy. They found a substantial number of biased entries, which were corrected.  

\subsubsection{Wordnet Affect}
Wordnet-Affect (WNA or WAL) is a hierarchical annotated Wordnet-based emotion lexicon comprising 289 noun synsets (a group of synonym words that express a notion). At least one emotion label has been assigned to each synset in Wordnet, which was picked from a set of labels in a specialized emotion hierarchy \cite{strapparava2004wordnet}.

\subsubsection{Alm's Fairy Tales}

Conducted in 2005, Cecilia Alm worked on emotion detection with 22 of Grimm’s tales, extracting sentences and then labeling them with 7 different emotions \cite{alm2005emotions}. These emotions were: Angry, Disgusted, Fearful, Happy, Sad, Positively Surprised, and Negatively Surprised, which is a modification of Ekman's model. From this, she was able to work on a total of 1,580 sentences with emotion labels that allowed her to help detect emotion in novels using supervised machine learning with the SNoW learning architecture.

\subsubsection{ISEAR}

The International Survey on Emotion Antecedents and Reactions dataset \cite{Scherer1994Evidence} is a collection of student responses when they were asked to report situations that occurred to them in which they had experienced 7 major emotions (Joy, Fear, Anger, Sadness, Disgust, Shame, and Guilt). Approximately 3,000 respondents in 37 countries responded detailing the way they had appraised the situation and their reaction towards it.



\subsubsection{EmoBank}

\newcite{buechel2017emobank} created the EmoBank dataset, a text corpus that was manually annotated based on the Valence-Arousal-Dominance scheme. This was done by collecting a large number of blogs, essays, news headlines, and other types of text to create a corpus that has around 10,000 English sentences. The dataset is annotated for Ekman's model of basic emotion (Anger, Sadness, Fear, Joy, Surprise, and Disgust).

\subsubsection{Emotion-Stimulus}

The emotion-stimulus dataset \cite{Ghazi2015Detecting} differs from the other datasets in that it is based on FrameNet’s {\it emotions-directed }frame while also noting what was the cause of the emotion being felt in a sentence. This dataset contains 820 sentences with both annotations while 1,594 sentences only have the emotion tags (Happiness, Sadness, Anger, Fear, Surprise, Disgust, and Shame, which is the Ekman model plus the Shame label).

\subsection{Language Resources for Animate Beings}

\newcite{bowman2012animacy} automatically annotated noun phrases based on a taxonomy of ten categories (Human, Org, Animal, Place, Time, etc). The corpus consists of around 600 transcribed dialogues from the parsed part of the Switchboard corpus~\cite{Calhoun2010switchboard}, which was coded by three undergraduate students from Stanford University. They then leveraged the results of \newcite{zaenen2004animacy} to distinguish which categories were  {\it animate} or {\it inanimate}. In the process of distinguishing and labeling both inanimate and animate their model achieved an $F_1$ 0.94 (versus a baseline model that labels only animate beings with an accuracy of 0.54).

\newcite{jahan2018animacy} noted issues with defining animacy as a word-level property, and instead proposed classifying the animacy of co-reference chains. In this work, they compiled a corpus using various resources such as Russian Folktales, Islamic Extremist Texts, Islamic Hadiths. In total, the corpus worked on consists of 142 texts, 156,154 tokens, 34,698 referring expressions, and 10,941 coreference chains. In her hybrid system using supervised Machine Learning and some written-rules, the system was able to achieve an $F_1$ of 0.88 in classifying the animacy of referring expressions, achieving state of the art performance. They further extended this work to the detection of characters in narrative \cite{jahan2020character}, annotating 30 texts from the Corpus of English Novels~\cite{desmet2008corpus} (among other works). We start from these 30 texts to construct the ABBE corpus described here.

\section{Annotation Scheme \& Process}
\label{sec:scheme}

We detail here the different parts and definitions of the ABBE annotation scheme, and describe how the corpus was annotated. 

\subsection{Annotation Scheme}

There are four components to annotation scheme: (1) the emotional expression span; (2) the emotion expressed in that span; (3) animate beings experiencing the identified emotions.

\subsubsection{Emotional Expression Span}

Annotators were first asked to identify spans of text that expressed emotions. They read the text and identified any emotional words or phrases. They were asked to identify the minimum span of contiguous tokens that covered all the emotional words in a single expression. Annotaters were allowed to reference WordNet \cite{fellbaum1998towards}  to help them disambiguate when words were being used in an emotional sense.

\ex. John was the [happiest] man alive

In this example, we mark only the token {\it happiest} as emotional, not including the article or the modified phrase {\it man alive}, as these are outside the minimum span of contiguous tokens that cover all the emotional words. In contrast:

\ex. Edna's death has filled me with immense [sorrow and dread] as I saw her life-less corpse.


The emotional span covers {\it sorrow and dread} since multiple tokens, separated by non-emotion words (i.e., {\it and}), are used to highlight the emotion occurring in this particular instance of the text which pertains to an article.

\subsubsection{Emotion Expressed}

Once the span was identified, annotators were asked to mark the span as to the emotion or emotions expressed, making this a multi-label classification task. As discussed, we used Plutchik’s primary  taxonomy of eight emotions: Joy, Trust, Fear, Surprise, Sadness, Anticipation, Anger, and Disgust. We provide descriptions of each category below.
    
\paragraph{Joy} A feeling of extreme gladness delight, or exaltation of the spirit arising from a sense of well-being.

\ex. ``The young man was fluent and \emo{gay}{Joy}, but he laughed louder than was natural in a person of polite breeding\ldots''

\paragraph{Sadness} An emotional state of unhappiness usually aroused by the loss of something that is highly valued

\ex. ``It was only forty pounds he needed,'' said the young man \emo{gloomily}{Sadness}.

\paragraph{Disgust} A strong aversion to something deemed revolting, or towards a person’s behavior deemed repugnant

\ex. From the whole tone of the young man's statement it was plain that he harbored very \emo{bitter
and contemptuous}{Disgust} thoughts about himself.

\paragraph{Trust} A reliance on or confidence in the dependability of something.

\ex. I \emo{confide}{Trust} in your abilities to get through all of this chaos.

\paragraph{Anger} An emotion characterized by tension and hostility arising from frustration.

\ex. All you do is \emo{anger}{Anger} me everyday.

\paragraph{Fear} An intense emotion aroused by the detection of imminent threat, involving an immediate alarm reaction.

\ex. ``Your Highness,'' said the Colonel, turning \emo{pale}{Fear}; ``let me ask you to consider the 
importance of your life, not only to your friends, but to the public interest.''

\paragraph{Anticipation} A looking forward to a future event or state, with an affective component.

\ex. He was eagerly waiting for it, \emo{expecting}{Anticipation} it to come for him.

\paragraph{Surprise} An emotion resulting from the violation of an expectation or the detection of novelty.

\ex. At its close Lady Theobald found herself in an utterly \emo{bewildered and thunderstruck}{Surprise} condition.

As mentioned, annotators were allowed to assign multiple emotion categories if a span expressed more than one emotion.

\ex. The poor man was \emo{jealous}{Anger, Disgust} of the rich kid winning the lottery. 


\ex. Remembering the tragic moment has instilled his soul with \emo{anger and sorrow}{Anger, Sadness} as he realized how corrupt the world around him was.

\ex. He could not contain himself, as he ran towards the entrance of the park \emo{gleefully looking forward}{Joy, Anticipation} to the adventures today will bring.


\subsubsection{Identifying the Emotion Experiencer}

With labeled emotional expression spans in hand, the last part of the annotation is to identify the animate being who is experiencing the emotion. To identify animate beings, we followed the same procedure as specified by \newcite{jahan2018animacy}. In particular, as we based ABBE off of the 30 selections from the Corpus of English Novels already annotated in that work so that we could check our annotations. The annotators looked for the closest referring expression for the relevant animate being based on their understanding of the text. If no animate being could be identified, the emotional span was dropped from the dataset. We found only minor variations between our annotations of animate beings and those of Jahan et al.

In the end, this sequence of steps resulted in annotations structured as follows:

\ex. \emo{His}{AB1} thoughts were both quiet and \emo{happy]}{Joy$\rightarrow$AB1}  His brief favour with the Duke he could not find it in his heart to mourn; with Joan to wife, and my Lord Foxham for a faithful patron, \emo{he}{AB2} looked most \emo{happily}{Joy$\rightarrow$AB2} upon the future; and in the past he found but little to regret.

In this short snippet of text there are two emotional experiences, the first being experienced by the person referred to by {\it His}, and the second by {\it he}. It just so happens that these two referring expressions are co-referent; we did not mark coreference explicitly in the dataset, but it can be extracted from Jahan et al.'s annotations of the same texts.

\subsection{Annotation Workflow}

As mentioned previously, the ABBE corpus is double-annotated. Two annotators (the first and second authors) performed the annotation. 
Each week, both annotators were given the same, specific collection of text to annotate. Once both annotators were finished, they met to review and adjudicate disagreements. 

\subsection{Agreement Measures}

Agreement on identification of the tokens that are part of emotion spans, before adjudication, was 0.933 $F_1$. Agreement on identification of animate beings, before adjudication, was 0.970 $F_1$. These two sets of judgements were then adjudicated, which allowed us to compute inter-annotator agreement for emotion assignment. 

The overall inter-annotator agreement on emotion assignment was 0.826 measured using Cohen's kappa ($\kappa$).  Cohen's kappa measures the agreement between two raters who each classify $N$ items into $C$ categories; here, there are $C=8$ different emotion categories. $\kappa$ is defined as $\frac {p_{o}-p_{e}}{1-p_{e}}$ where $p_o$ is the relative observed agreement. Assuming that emotion sets $L_1$ and $L_2$ represent the multi-emotion labels given by raters 1 and 2 respectively, the relative observed agreement is calculated as:

\begin{equation}
p_o = \frac{|(L_1\cap L_2)\cup(\overline{L_1}\cap\overline{L_2})|}{8}
\end{equation}

where 8 is the number of emotion labels. $p_e$ is the hypothetical probability of chance agreement, using the observed data to calculate the probabilities of each observer randomly seeing each category. We used the following formula to calculate $p_e$:

\begin{equation} 
p_e = \frac{\sum\limits_{k}N_{1k}N_{2k}}{N^2}
\end{equation}

where $N_{1k}$ and $N_{2k}$ are the number of times that raters 1 and 2 predicted category $k$.


\section{Selected Texts}
\label{sec:texts}


ABBE comprises 30 selections from the Corpus of English Novels, a 25-million-word corpus which consists of 292 novels by 25 different novelists \cite{desmet2008corpus}. These novels were written between 1881 and 1922 and the corpus was designed to track short-term language changes and comparing usage across a certain generation of authors. The purpose of selecting novels for ABBE is to take advantage of the relative density and variety of reported emotion due to the nature of the genre and a large number of different characters, environments, situations, and writing styles provided by the different novels from different writers. The key counts for the corpus are given in Table~\ref{tab:corpus}, while the specific selections included in ABBE are shown in Table~\ref{tab:texts}

\begin{table}[]
    \centering
    \begin{tabular}{ll}
    \toprule
    \bf Feature      & \bf Count \\
    \midrule
    \# of Texts      & 30 \\
    \# Tokens        & 134,513 \\
    \# Emotional Animate Beings & 2,227 \\
    \# Emotion Spans & 2,010 \\
    \midrule
    \# Joy labels          & 735 \\
    \# Sadness labels      & 678 \\
    \# Disgust labels      & 359 \\
    \# Trust labels        & 471 \\
    \# Anger labels        & 408 \\
    \# Fear labels         & 540 \\
    \# Anticipation labels & 649 \\
    \# Surprise labels     & 409 \\
    \bottomrule
    \end{tabular}
    \caption{Key Counts for the ABBE Corpus}
    \label{tab:corpus}
\end{table}


\begin{table*}[t]
\centering
\begin{tabular}{llll}
\toprule
Novels        & Name Of Story                                               & Author                          & Chapters \\
\midrule
1             & A Fair Barbarian                                            & Frances Hodgson Burnett         & 25-26    \\
2             & Milly and Olly                                              & Mary Augusta Ward               & 10       \\
3             & Treasure Island                                             & Robert Louis Stevenson          & 11       \\
4             & Mr. Isaacs                                                  & Francis Marion Crawford         & 14       \\
5             & The Suicide Club                                            & Robert Louis Stevenson          & 1      \\
6             & Doctor Claudius: A True Story                               & Francis Marion Crawford         & 20       \\
7             & A Roman Singer                                              & Francis Marion Crawford         & 24       \\
8             & Miss Bretherton                                             & Mary Augusta Ward               & 7        \\
9             & Philistia                                                   & Charles Grant Blairfindie Allen & 2        \\
10            & The Black Arrow: A Tale of the Two Roses                    & Robert Louis Stevenson          & 7        \\
11            & The Unclassed                                               & George Robert Gissing           & 38       \\
12            & A Mummer's Wife                                             & George Augustus Moore           & 30       \\
13            & King Solomon's Mines                                        & Henry Rider Haggard             & 20       \\
14            & Little Lord Fauntleroy                                      & Frances Hodgson Burnett         & 1        \\
15            & Prince Otto                                                 & Robert Louis Stevenson          & 4        \\
16            & The Children of the King                                    & Francis Marion Crawford         & 12       \\
17            & The Shadow of a Crime                                       & Thomas Henry Hall Caine         & 51       \\
18            & Zoroaster                                                   & Francis Marion Crawford         & 20       \\
19            & A Tale of a Lonely Parish                                   & Francis Marion Crawford         & 24       \\
20            & Demos                                                       & George Robert Gissing           & 36       \\
21            & On Being in Love                                            & Jerome Klapka Jerome            & 2        \\
22            & Kidnapped                                                   & Robert Louis Stevenson          & 30       \\
23            & Muslin                                                      & George Augustus Moore           & 29       \\
24            & A Romance Of Two Worlds                                     & Marie Corelli                   & 14       \\
25            & Strange Case of Dr. Jekyll and Mr. Hyde                     & Robert Louis Stevenson          & 1        \\
26            & Vendetta!: Or The Story of One Forgotten, a Novel, Volume 1 & Marie Corelli                   & 1        \\
27            & A Mere Accident                                             & George Augustus Moore           & 9        \\
28            & Marzio's Crucifix, Volume 1                                 & Francis Marion Crawford         & 11       \\
29            & A Little Princess                                           & Frances Hodgson Burnett         & 18       \\
30            & Saracinesca                                                 & Francis Marion Crawford         & 34       \\
\bottomrule
\end{tabular}
\caption{CEN 30 Selected Texts}
\label{tab:texts}
\end{table*}

\section{Difficult and Interesting Cases}
\label{sec:edgecases}


\subsection{Multiple Conflicting Emotions}

Naturally it is possible for an animate being to feel different, even conflicting, emotions over time. There is no problem when these emotion mentions are separated in the text, but in certain cases the text presents these emotions as being attached to the same referring expression. In these cases we do not provide any distinguishing information, and just annotation the emotion mentions normally:

\ex. \emo{Henry}{AB1} was \emo{upset}{Sadness$\to$AB1} at his grade for his midterm, but this turned to \emo{joy}{Joy$\to$AB1} when it turned out to actually be a passing grade. \label{ex:henry}

\ex. \emo{Anger}{Anger$\to$AB1} filled \emo{John’s}{AB1} veins, which soon became tears of \emo{sorrow}{Sadness$\to$AB1}\ldots

In both examples, the relevant animate being moves from one emotion to another, but the emotions are attached to the same referring expression.

\subsection{Sets of Animate Beings}

It is not uncommon for a single emotional span to be attributable to several animate beings at once. In the case of separable mentions, the annotation points to each individual animate being. For example:

\ex. \emo{Jack}{AB1} was afraid of \emo{falling}{Fear$\to$AB1,AB2}, and so was \emo{Jill}{AB2}.

In cases where the animate beings are indicated by a single phrase (such as a conjunctive noun phrase, or a plural pronoun), the phrase is marked as the relevant animate being:

\ex. \emo{Jack and Jill}{AB1} were \emo{scared}{Fear$\to$AB1} of falling.

\ex. \emo{They}{AB1} were \emo{scared}{Fear$\to$AB1} of falling. 

In these cases, the emotion is attributed to the set that comprises {\it Jack and Jill} (or {\it they}).

\subsection{Emotion vs. Action}

There is a difference between emotion and action; while there are many actions which imply the agent is or will experience certain emotions, we don't mark an emotion unless it is necessarily part of the semantics of the action, or the emotion is explicitly mentioned. Compare the following:

\ex. I went to Disneyland. (no emotion)
\ex. \emo{I}{AB1} was \emo{ecstatic}{Joy$\to$AB1} to go to Disneyland. 







\subsection{Emotion vs. Mood}

There are also cases where an emotion is not present but is generalized, not isolated in time, and cannot be assigned to a specific animate being or set of animate beings. These we call {\it moods}, and we do not mark them. For example:
    
\ex. \underline{Dread} fills the city late at night.
    
In this sentence, Dread is clearly affective, but there is no one specifically mentioned as experiencing the emotion. 


\section{Contributions}
\label{sec:contributions}

Our contributions in this paper are threefold. First, we note the importance of the experiencer to the idea of emotion, and point out that all prior corpora of annotation emotion omit this information. Second, we define an annotation scheme that captures the experiencer of an emotion (an animate being), based on prior work on emotion classification and animate being detection. Finally, we provide the ABBE corpus, a collection of 30 texts (134.5k tokens) annotated for 2,010 emotion spans associated with 2,227 animate beings, annotated with excellent inter-annotator agreement.  We release the ABBE data for other researchers to use in their work\footnote{\codecite}.


\bibliographystyle{lrec}
\bibliography{paper}

\end{document}